\documentclass[letterpaper, 10 pt, conference]{ieeeconf}
\IEEEoverridecommandlockouts
\usepackage{cite}
\usepackage{amsmath,amssymb,amsfonts}
\usepackage[noend]{algorithmic}
\usepackage{graphicx} 
\usepackage{color}
\usepackage{xcolor}
\usepackage{subcaption}
\usepackage{hyperref}
\usepackage{float}
\usepackage{diagbox}
\usepackage{comment}
\usepackage{svg}
\usepackage{import}

\usepackage{algorithm}
\usepackage{mathtools}

\def\BibTeX{{\rm B\kern-.05em{\sc i\kern-.025em b}\kern-.08em
    T\kern-.1667em\lower.7ex\hbox{E}\kern-.125emX}}
\begin{document}

\title{\LARGE \bf N(CO)$^2$: Neural Combinatorial Optimization with Chance Constraints to Solve Stochastic Orienteering}


\author{Anas Saeed \qquad Marcos Abel Zuzu\'{a}rregui \qquad Stefano Carpin%
\thanks{
The authors are with the Department of Computer Science and Engineering, University of California, Merced, CA, USA.
This work is partially supported by the IoT4Ag Engineering Research Center funded by the U.S. National Science Foundation (NSF) under NSF Cooperative Agreement Number EEC-1941529, and by the US Department of Agriculture
under award
\#2021-67022-33452 (National Robotics Initiative). Any opinions, findings, conclusions, or recommendations expressed in this publication are those of the author(s) and do not necessarily reflect the view of the National Science Foundation or the US Department of Agriculture.}
}

\maketitle

\begin{abstract} 
Neural combinatorial optimization (NCO) offers a promising alternative to traditional heuristic-based methods for solving complex graph optimization problems by proposing to learn heuristics through data. This class of problems frequently arises in automation, as it can be used to model a variety of applications. While NCO has been extensively studied for deterministic combinatorial optimization problems, there are only a few works that aim to solve stochastic combinatorial optimization problems.
In this work, we present N(CO)$^2$: Neural Combinatorial Optimization with Chance cOnstraints to solve the Stochastic Orienteering Problem (SOP) without the use of hand-crafted heuristics.
By integrating a reinforcement learning (RL) framework, the model optimizes path selection under uncertainty, effectively balancing exploration and exploitation.
Empirical results demonstrate that our method generalizes well across diverse SOP instances, achieving competitive performance compared to the state-of-the-art mixed-integer linear program (MILP) for the task.
The proposed approach reduces human effort in heuristic design while enabling adaptive and efficient decision-making in uncertain environments.

\end{abstract}

\section{Introduction} 

\begin{figure}[tb!]
    \centering
    \includegraphics[width=0.8\linewidth]{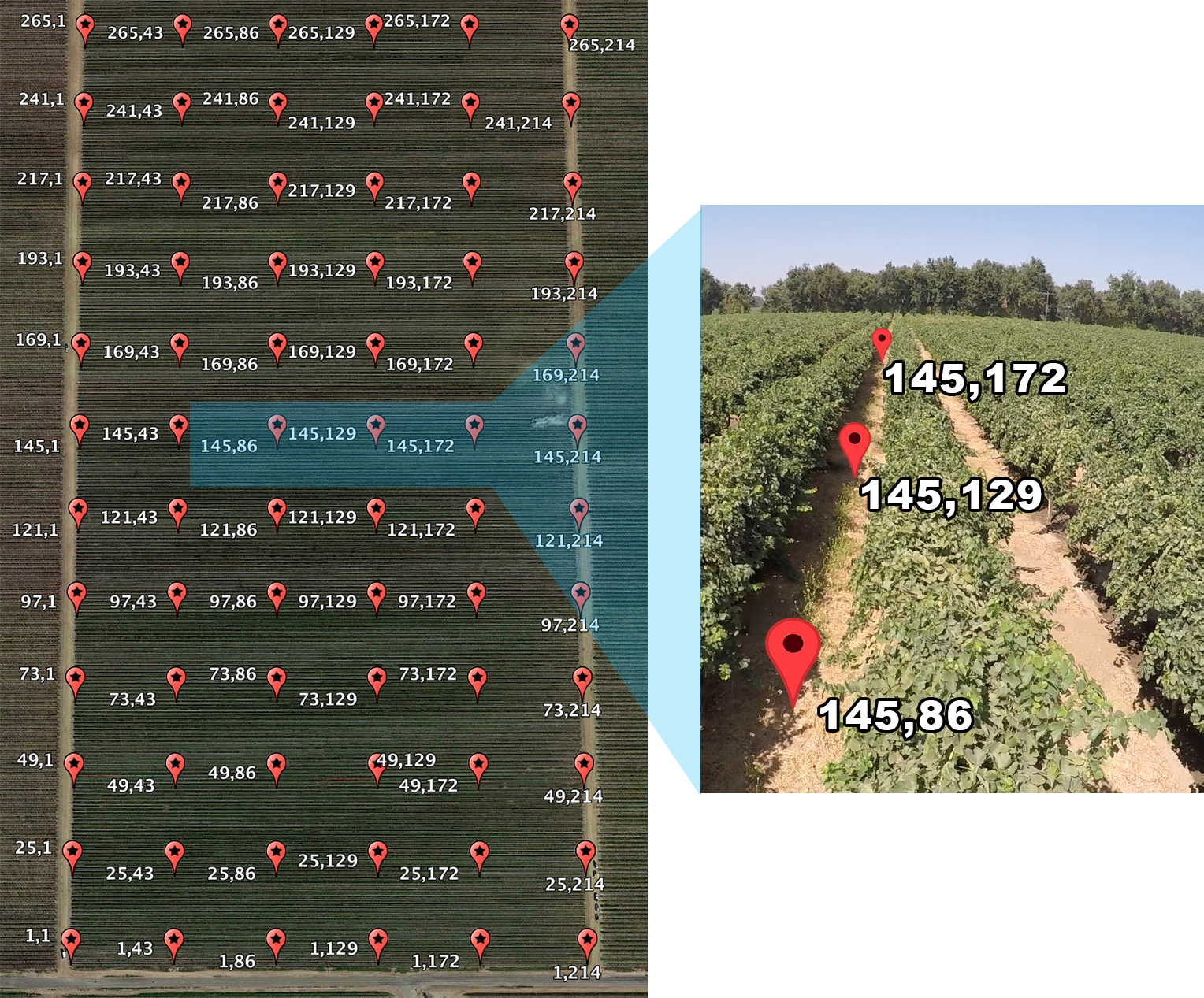}
   \caption{An example of a problem that can be modeled using the 
    orienteering COP. 
    Figure taken from \cite{CarpinTASE2024}.}
    \label{fig:motivation}
    \vspace{-5mm}
\end{figure}

Combinatorial optimization problems (COPs) can often naturally be reduced to graph representations.
Many COPs are NP-hard, leading to the development of various heuristics to approximate optimal solutions within reasonable time constraints.
However, this results in a set of solutions that are hand-crafted to solve only a single problem within COPs.
The question we ask ourselves is: \emph{what if we could create a general architecture using reinforcement learning (RL) and neural combinatorial optimization (NCO) fundamentals to eliminate the need to recreate heuristics for solving instances of COPs?}
To illustrate our motivations, we select one COP of interest based on our ongoing research in precision agriculture and path planning \cite{CarpinTASE2024, zuzuarregui_solving_2024, CarpinLLMSubmitted}: orienteering.
This APX-hard problem \cite{Golden1987} is related to the Traveling Salesman Problem (TSP) and is formulated as a graph optimization problem.
In precision agriculture, models of this type arise when considering problems where a robot has finite energy and is used to perform a set of tasks in the field (e.g., collecting a set of soil moisture measurements). 
Since not all locations are equally informative for sampling, and the robot cannot exhaustively collect samples at all possible locations, the orienteering abstraction can be used to solve this resource allocation problem (see Figure \ref{fig:motivation}). 
While most of the literature aims to address deterministic COPs, this paper focuses on the stochastic variant of orienteering.
This version, known as the stochastic orienteering problem (SOP), has received far less attention due to the difficulty of modeling stochasticity.
Stochasticity in SOPs is realized as uncertainty in the cost $c_i$ incurred while traversing $e_i$ in an attempt to reach the goal vertex $v_g$, which is only realized at runtime.
This variant of orienteering is a theoretical formulation of problems that arise not only in precision agriculture but also in real-world navigational scenarios.
Problems such as path planning \cite{zuzuarregui_solving_2024}, logistics, and even ridesharing \cite{Martin2021} can be reduced to this graph optimization problem.

Our previous works implement versions of Monte Carlo Tree Search (MCTS) aimed at solving the SOP with and without neural networks.
However, in both cases, they require a manually developed heuristic to either solve the problem \cite{CarpinTASE2024} or train a network to help solve the problem \cite{zuzuarregui_solving_2024}.
In crafting a heuristic to solve a problem, we are limited in the strength and flexibility of the solution due to the heuristic typically disallowing effective exploration.
Moreover, writing an effective heuristic for a problem with stochasticity is much more difficult than writing one for a deterministic problem.
We now aim to capture a balance between the speed of solutions generated in \cite{zuzuarregui_solving_2024} and the power and generalizability of \cite{CarpinTASE2024}, without manually formulating a heuristic well suited for the problem.
In this work, we investigate the use of a random walk algorithm to generate effective training data for an edge-augmented graph transformer (EGT) model, intending to capture the likelihood of edge selection as part of an SOP path solution.
The main contributions of our paper are as follows:

\begin{itemize}
  \item We propose 
  N(CO)$^2$: Neural Combinatorial Optimization with Chance cOnstraints,
  a neural network-based heuristic learning algorithm for stochastic combinatorial optimization problems using chance constraints.
  \item We introduce a scalable solution construction algorithm for stochastic orienteering using our learned heuristic.
  \item We demonstrate that our proposed combination of learned heuristics and solution construction produces solutions for SOPCC on par with SOTA MILP solutions, at faster speeds.
\end{itemize}

The rest of the paper is organized as follows.
Related work is presented in Section \ref{sec:sota}.
The SOP is formalized in Section \ref{sec:background}.
In that section, we also discuss the RL formulation and our foundational model used to train our heuristic network.
Section \ref{sec:methodology} describes the generated heuristic model, as well as the data, feature sets, and RL framework.
Extensive simulations detailing our findings are given in Section \ref{sec:results}, with conclusions and future work discussed in Section \ref{sec:conclusions}.
\section{Related Literature} \label{sec:sota}

\subsection{Neural Combinatorial Optimization}

Neural Combinatorial Optimization (NCO) \cite{bello2016neural} is an emerging paradigm for solving a variety of NP-hard problems on graphs using neural networks. NCO algorithms are able to learn heuristics for COPs using supervised training examples \cite{joshi2022learning, sun2023difusco}, from scratch with reinforcement learning \cite{bello2016neural, kool2018attention, qiu2022dimes, grinsztajn2023winner}, or using unsupervised learning \cite{min2023unsupervised}.

Our method is based on non-autoregressive (NAR) techniques for NCO \cite{joshi2022learning, qiu2022dimes, min2023unsupervised}, which formulate the problem as an edge prediction task. NAR techniques generate the solution in one shot by predicting an edge heatmap that defines how likely a given edge will occur in the optimal solution. Using the edge heatmap, solutions to the COP can be decoded using construction-based algorithms.

Recent research has applied NCO to stochastic and constrained CO problems, such as portfolio optimization \cite{dai2023dominance, wang2023linsatnet} and stochastic routing problems \cite{xiao2024neural, tong2021uscosolver, smit2024sjsp}. However, to our knowledge, N(CO)$^2$ is the first algorithm to apply a reinforcement learning-based approach to train neural networks for the Stochastic Orienteering Problem.

\subsection{Optimization Algorithms for Orienteering}

Our recent works \cite{CarpinTASE2024, zuzuarregui_solving_2024} use MCTS as an online approach to solve the SOP, selecting vertices dynamically based on the remaining budget.
\cite{CarpinTASE2024} uses classical computation, and \cite{zuzuarregui_solving_2024} implements a graph neural network (GNN) trained on the heuristic developed in \cite{CarpinTASE2024}.
Both are competitive, but \cite{CarpinTASE2024} is not real-time due to large state space approximation, and \cite{zuzuarregui_solving_2024} struggles to generalize failure probability estimates.
An exact approach in \cite{MILP} constructs a mixed-integer linear program (MILP).
However, this is an offline approach that differs by orders of magnitude in solution generation timing.
Additionally, studies on deterministic TSP and orienteering have demonstrated the effectiveness of message-passing frameworks and graph attention \cite{DBLP:journals/corr/DaiKZDS17, 9109309, LIU202446}.
While GNN-based methods have been explored for orienteering, no prior work has used machine learning to effectively generalize SOP solutions at near-optimal levels.


\section{Background} \label{sec:background} 

In this section, we formally introduce the stochastic orienteering problem with chance constraints (SOPCC). 
We then outline other variants of MCTS that inspired this solution and their applications to SOPCC.

\subsection{Stochastic Orienteering Problem with Chance Constraints}
The classical orienteering problem is formulated as follows.  
Consider a weighted graph \( G = (V, E) \) with \( n \) vertices, where \( V \) denotes the set of vertices and \( E \) represents the set of edges. Without loss of generality, we assume that \( G \) is a complete graph, i.e., \( E = V \times V \).  
We define \( r : V \rightarrow \mathbb{R}_{+} \) as the reward function that assigns a positive reward to each vertex, and let \( c : E \rightarrow \mathbb{R}_{+} \) be the cost function that assigns a positive cost to each edge.  
Let \( v_s, v_g \in V \) be the designated start and goal vertices, respectively, with a fixed budget \( B > 0 \).  
We allow both the cases where \( v_s = v_g \) or \( v_s \neq v_g \).  
For a given path \( \mathcal{P} \) in \( G \), \( R(\mathcal{P}) \) is the total reward accumulated from the vertices along \( \mathcal{P} \), and \( C(\mathcal{P}) \) is the total cost of the edges in \( \mathcal{P} \).  
The orienteering problem asks to solve the following constrained optimization problem:
\[
\mathcal{P}^* = \arg \max_{\mathcal{P} \in \Pi} R(\mathcal{P}) \quad \text{s.t.} \quad C(\mathcal{P}^*) \leq B,
\]
where \( \Pi \) represents the set of paths in \( G \) that start at \( v_s \) and end at \( v_g \) and never revisit the same vertex  
(it is trivial to show that allowing paths to visit the same vertex twice or more does not yield better solutions).  
Given that \( G \) is assumed to be complete, restricting \( \Pi \) in this way does not impose additional limits.  
In the stochastic variant, the cost associated with each edge follows a continuous random variable with a known probability density function (PDF) that has strictly positive support.  
Specifically, for each edge \( e_i \in E \), the cost \( c(e_i) \) is sampled from a distribution \( d(e_i) \), which represents the random variable modeling the traversal cost of the edge.  
Consequently, the total path cost \( C(\mathcal{P}) \) is itself a random variable.  
This necessitates expressing the budget constraint probabilistically using a chance constraint, formally defined as follows.

\begin{quote} 
Given the notation introduced above, let $0 < P_f < 1$ represent an assigned maximum failure probability threshold. 
The SOPCC seeks to solve the following optimization problem:
\begin{align}
\mathcal{P}^* &= \arg \max_{\mathcal{P} \in \Pi} R(\mathcal{P}) \nonumber \\
\text{s.t.}~ &\Pr[C(\mathcal{P}^*) > B] \leq P_f \label{eq:cc}
\end{align}
\end{quote}

This problem formulation captures the goal of maximizing the collected reward while ensuring that the probability of exceeding the budget remains within an acceptable bound.  
Due to the stochastic nature of \( C(\mathcal{P}) \), the constraint can only be satisfied probabilistically, leading to the introduction of the chance constraint (\ref{eq:cc}).

\subsection{Sample Average Approximation}
A method for computing the estimated failure chance \( F \) is Sample Average Approximation (SAA) \cite{SAApaper}.  
With this approach, \( N \) independent, identically distributed samples are taken of a random variable \( \xi \).  
These samples can then be used to approximate the probabilistic constraint included in Eq.~\eqref{eq:cc} and to satisfy the chance constraint in the SOPCC problem formulation.  
To do so, we draw \( N \) samples from a PDF to approximate the estimated probability of failure of a given path \( \mathcal{P} \in \Pi \), shown in \eqref{eq:saa} as \( \hat{p}_N(F) \).  
This helps to identify feasible paths that do not satisfy our constraint.  
Formally, we define this SAA approach with respect to the SOP with chance constraints as follows:
\begin{align}
\hat{p}_N(F) = \frac{1}{N}\sum_{i=1}^N \mathbb{I}\big(C(\mathcal{P}, \xi_i) > B \big) \label{eq:saa}
\end{align}
where \( \mathbb{I} \) is the indicator function that is equal to 1 if its argument is true and 0 otherwise.  
The assumption, as per the law of large numbers, is that the larger \( N \) is, the closer \( \hat{p}_N(F) \) approaches the true value \( p(F) \).

\section{Methodology} \label{sec:methodology}

This section introduces the main contributions of our paper.

\subsection{Heuristic Based Solution to SOPCC}
As pointed out in \cite{CarpinTASE2024}, one can build a solution to the SOPCC problem by iteratively building a path \( \mathcal{P} \) adding one vertex at a time.  
This strategy is indeed used in \cite{CarpinTASE2024} to implement the rollout stage in the MCTS construction and explore the space of possible solutions.  
This approach can be seen as a greedy algorithm, inasmuch as it never reconsiders past choices: once a vertex is added to the path, it will remain there.  
A critical aspect of this approach is deciding which vertex should be added next. This is typically done using a heuristic, and as we pointed out in the introduction, the choice of the heuristic may have a large impact on the performance of algorithms like MCTS that repeatedly use heuristics to build better solutions.  
In designing a heuristic for the SOPCC problem (or, more generally, for problems with chance constraints), one additional challenge is that the heuristic should consider the constraints, i.e., greedy choices violating the constraint should be rejected.  
We start by providing a high-level algorithm that shows how, with a given heuristic \( H \) that for the time being can be considered a \emph{black box}, a path \( \mathcal{P} \) solving the SOPCC problem can be iteratively computed. Algorithm \ref{alg:walk} sketches this approach.  
Besides the parameters defining the SOPCC instance, the algorithm accepts as input the heuristic to be used \( H \) and \( S \), the number of samples to be used while applying SAA to estimate the failure probability.  
Inside Algorithm \ref{alg:walk} we keep two variables to respect the constraint. \( F_{\text{mask}} \) is the set of vertices that have been determined to violate the chance constraint and should not be considered when expanding the path.  
Importantly, this set depends on the last vertex added to the path, so every time a new vertex is added (line 10), \( F_{\text{mask}} \) is reset (line 12).  
The other variable is \( B_{\text{samples}} \), which is a vector in \( \mathbb{R}^S \) that stores the residual budget.  
That is to say, at each iteration \( B_{\text{samples}} \) is a vector of \( S \) samples modeling the remaining budget after having constructed the current partial path \( \mathcal{P} \).  
\( B_{\text{samples}} \) is initialized at line 1 with \( S \) copies of \( B \) and is then updated every time a vertex is added to the partial path \( \mathcal{P} \), either in line 11 or in line 17.  
The algorithm uses two functions, namely \texttt{selectAction} and \texttt{sCosts}. \texttt{selectAction} uses the heuristic \( H \) to select a vertex to add to the path \( \mathcal{P} \) under the assumption that \( v_c \) was the last vertex added. In doing so, it ensures that vertices in \( F_{\text{mask}} \) are not considered because they would violate the constraint.  
\texttt{sCosts}(\( v_i, v_j, S \)) (short for sample costs) is a function that uses the PDF to generate \( S \) samples of the cost of moving from \( v_i \) to \( v_j \) along the edge that connects them.  
In line 9, we use the SAA method to accept or reject the proposed vertex \( v_n \) based on whether it violates the probabilistic constraint or not.

\begin{algorithm}[hbt!]
\caption{Solution Construction with Heuristic}
\begin{algorithmic}[1]
\REQUIRE heuristic $H$, graph $G$, starting node $v_s$, goal node $v_g$, initial budget $B_i$, failure constraint $P_f$, samples $S$
\ENSURE path $\mathcal{P}$, residual sampled budget $B_{samples}$
\STATE $B_{samples}$ $\gets$  $S$ copies of $B$
\STATE $F_{mask} \gets \{\}$ \COMMENT{Adjacent nodes that violate the failure constraint}
\STATE $v_c \gets v_s$
\STATE $\mathcal{P} \gets \{v_s\}$
\WHILE{true}
  \STATE $v_{n} \gets \text{selectAction}(v_c, H,\mathcal{P}, F_{mask})$
  \IF{$v_n \neq v_g$}

    \STATE $cost \leftarrow\text{sCosts}(v_c, v_n, S) + \text{sCosts}(v_n, v_g, S)$ 

    \IF{$\Pr[(cost > B_{samples}))]\leq P_f$}
        \STATE append $v_n$ to $\mathcal{P}$
        \STATE $B_{samples} \gets B_{samples} - \text{sCosts}(v_c, v_n, S)$
        \STATE $F_{mask} \gets \{\}$
    \ELSE
        \STATE append $v_n$ to $F_{mask}$
    \ENDIF
  \ELSE
    \STATE append $v_g$ to $\mathcal{P}$
    \STATE $B_{samples} \gets B_{samples} - \text{sCosts}(v_c, v_g, S)$
    \STATE break
  \ENDIF
\ENDWHILE
\RETURN $\mathcal{P}$, $B_{samples}$
\end{algorithmic}
\label{alg:walk}
\end{algorithm}

Given the dependency of \texttt{selectAction} on the heuristic \( H \), we next discuss how \( H \) can be represented and learned from data.

\subsection{Solution Space with a Parameterized Heuristic}
In this section, we lay the foundation for the probabilistic heuristic \( H \) used in \texttt{selectAction}. 
The heuristic is probabilistic, i.e., when queried it assigns to each vertex a probability, and then it samples from such a probability distribution. 
For an SOPCC instance where the graph \( G \) has \( n \) nodes, our heuristic \( H \) is defined by a matrix\footnote{We intentionally use the same symbol for the heuristic and the matrix, because the matrix defines the heuristic.} \( H \in \mathbb{R}^{n \times n} \), which, after a softmax operation, maps each edge \( e_{ij} \) to a probability \( H_{ij} = \Pr[e_{ij} \in \mathcal{P}] \). The idea is that \emph{good} edges should receive higher probabilities. In this context, \( H \) is also referred to as a heatmap \cite{joshi2022learning, qiu2022dimes}. 

In essence, the \((i,j)\) entry defines the probability that the edge from \( v_i \) to \( v_j \) should be added to the path, and \( H \) is parametrized by a vector of parameters \(\theta\) that will be learned from data. To make this dependency explicit, in some of the following formulas, we will therefore write \( H_{\theta} \).

Following \cite{qiu2022dimes}, for a given matrix \( H \), we first apply the following transformation:
\[
H(i,j) = \begin{cases} 
    H(i,j) & \text{if } v_j \notin F_{mask} \wedge v_j \notin \mathcal{P} \\
    0  & \text{otherwise}
\end{cases}
\]
i.e., we zero out the entries associated with vertices that would violate the chance constraint. These are included in \( F_{mask} \), as per Algorithm \ref{alg:walk}, or are already in the partial path \(\mathcal{P}\). From this matrix, we then assign probabilities to the vertices using a softmax operation. 

In the following, \(\Pr[v_j | v_i, \mathcal{P}]\) is the probability of adding vertex \( v_j \) to a partial path \(\mathcal{P}\) whose last vertex is \( v_i \):
\[
\Pr[v_j | v_i, \mathcal{P}] \coloneqq \frac{\exp(H(i,j))}{\sum_{k=1}^n \exp(H(i,k))}
\]

Finally, for the learning algorithm described later, it is useful to introduce the following quantity, which is the probability that a certain path \(\mathcal{P}\) will be produced by the heuristic \( H \):
\[
\Pr[\mathcal{P} | H] = \prod_{k=1}^{K-1} \Pr[v_{k+1} | v_k, \mathcal{P}]
\]
where \( K \) is the number of vertices in the path (recall that each path starts with vertex \( v_1 = v_s \) as per Algorithm \ref{alg:walk}).

\subsection{Heuristic Improvement with Reinforcement Learning}
Having defined the structure of the heuristic function through
a heatmap matrix parametrized by a vector $\theta$, the next question, then, is how $\theta$ can be learned.
To this end, we embrace a reinforcement learning approach,
and more precisely, we leverage the classic REINFORCE algorithm.
Reinforcement learning needs a reward signal, which in our case
scores a path $\mathcal{P}$ produced by Algorithm \ref{alg:walk}.
The challenge is that just seeking a path maximizing reward is not 
sufficient, because we need to also consider the failure probability.
Therefore, we introduce the following function to assign
a reward to a path $\mathcal{P}$:

\begin{equation}
f_{s}(\mathcal{P}, \hat{F}_{\mathcal{P}}, P_f) = R(\mathcal{P}) \cdot \bigl(1-\max(0, \hat{F}_{\mathcal{P}} - P_f)\bigr)
\label{eq:score}
\end{equation}

where $\hat{F}_{\mathcal{P}}$ is the estimated failure of path $\mathcal{P}$
computed using the SAA algorithm on the $B_{samples}$ vector
returned by Algorithm \ref{alg:walk}.
This reward function penalizes the reward of solutions where $\hat{F}_{\mathcal{P}}$ exceeds $P_f$, but does not affect solutions which satisfy $P_f$. It should be noted that in the case where an infeasible path contains significantly more reward than all feasible paths, $f_s$ will in principle favor that solution. In practice, however, this is not an issue because 
in Algorithm \ref{alg:walk} such paths are unlikely to be generated 
thanks to the use of SAA inside the algorithm itself.

Our training objective is formulated as follows:
\begin{equation}
J(\theta) = \mathbb{E}_{\mathcal{P} \sim H_{\theta}} [f_s(\mathcal{P})]
\label{eq:rl_objective}
\end{equation}

For this objective function we compute the policy gradient with the REINFORCE \cite{Williams1992} based gradient estimation following \cite{qiu2022dimes}:
\begin{equation}
\nabla_{\theta} J(\theta) = \mathbb{E}_{\mathcal{P} \sim H_{\theta}} \left[(f_s(\mathcal{P}) - b) \times \nabla_{\theta} \log \Pr[\mathcal{P} | H] \right]
\label{eq:reinforce_baseline}
\end{equation}
where $b$ is the baseline function representing the expected cost of the problem instance. In our experiments, we use the baseline function introduced in \cite{kool2019buy}, which approximates the expected cost by averaging the score of $K$ sampled solutions. 

It should be noted that using \eqref{eq:reinforce_baseline} does not directly optimize the sampled population towards the best path, but rather optimizes the average performance of the population. To this end, we compare against the population-based REINFORCE objective introduced in the Poppy \cite{grinsztajn2023winner} paper:
\begin{align}
\nabla_{\theta} J_{pop}(\theta) &= \mathbb{E}_{\mathcal{P} \sim H_{\theta}} \left[(f_s(\mathcal{P}^*_i) - f_s(\mathcal{P}^{**}_i)) \right. \label{eq:reinforce_population} \\
& \qquad \times \left. \nabla_{\theta} \log \Pr[\mathcal{P}^*_i | H] \right] \notag
\end{align}
where in a population of $K$ solutions sampled using $H_{\theta}$, $\mathcal{P}^*_i$ is the solution that has the highest $f_s$ and $\mathcal{P}^{**}_i$ is the solution that has the second highest $f_s$. While \eqref{eq:reinforce_baseline} allows the model to use all $K$ sampled paths to update the model, the population-based objective only updates the model with the best path $\mathcal{P}^*_i$. 

In Section \ref{sec:results} we will compare the results of both strategies.
Based on the two sampling-based approaches mentioned above, we describe the general algorithm we use to train N(CO)$^2$ in Algorithm \ref{alg:pretraining}. For each graph instance, we sample $K$ solutions using our generated heuristic $H_{\theta}$, and use those solutions to optimize our objective function \eqref{eq:rl_objective}. We use the AdamW \cite{loshchilov2018adamw} optimizer to update the model parameters. To stabilize training, we clip the gradient values between $[-0.5, 0.5]$.

\begin{algorithm}[hbt!]
\caption{RL Pretraining Algorithm for SOPCC}\label{alg:pretraining}
\begin{algorithmic}[1]
\REQUIRE distribution over problem instances $\mathcal{D}$, batch size $B_S$, number of training steps $T$, number of solutions per step $K$, number of samples $S$
\ENSURE learned network parameters $\theta$
\STATE randomly initialize network parameters $\theta$
\FOR{$t=1$ to $T$}
    \STATE $\rho_i \gets \text{SampleProblem}(\mathcal{D})$ for $i\in\{1,\dots,B_S\}$
     \STATE Build $H_{\theta_i}$ from current instance $\rho_i$
    \STATE Sample K solutions for $\rho_i$ using Algorithm \ref{alg:walk} for $i\in\{1,\dots,B_S\}$ using $H_{\theta_i}$
    \STATE Compute $\nabla_{\theta}J(\theta\mid\rho_i)$ using Eq.~ \eqref{eq:reinforce_baseline} (or Eq.~\eqref{eq:reinforce_population}
    \STATE $\theta \gets \text{AdamW}(\theta, -\nabla_{\theta}J)$
\ENDFOR
\RETURN $\theta$
\end{algorithmic}
\end{algorithm}


\begin{figure*}[t!]
    \centering
    \includegraphics[width=500pt]{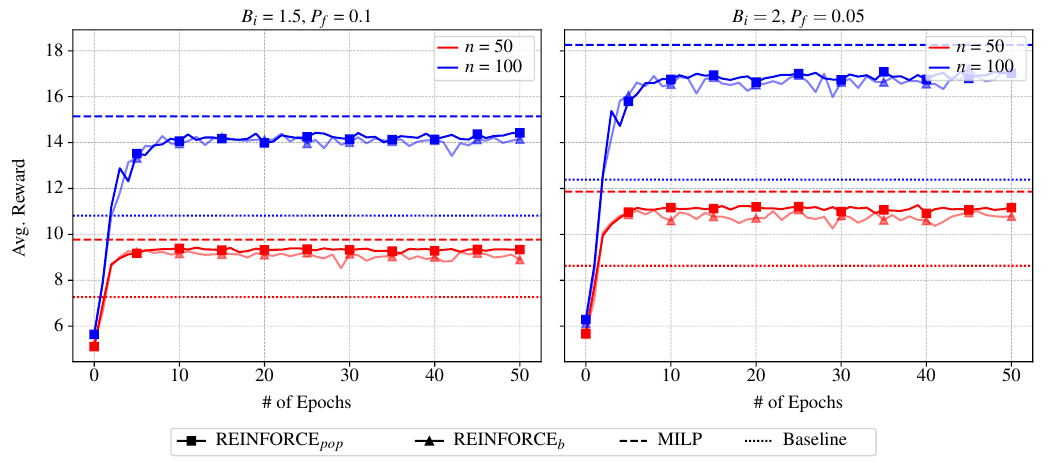}
    \caption{Comparison of training performance of the model using Eq.~\eqref{eq:reinforce_baseline} and \eqref{eq:reinforce_population} on four evaluation datasets. For each evaluation dataset, the reward on the vertical axis is averaged across all graph instances.}
    \label{fig:eval}
\end{figure*}

\subsection{Per-instance Finetuning with Active Search}

We can further improve the performance of our model during inference using active search \cite{bello2016neural, qiu2022dimes}, which uses the same RL-based objective 
in Eq.~\eqref{eq:rl_objective} to directly optimize our model on the target instance. We describe the active search algorithm for N(CO)$^2$ in Algorithm \ref{alg:activesearch}. Since only the generated heuristic $H_\theta$ is used for solution construction, we can directly update $H_\theta$ while keeping the network parameters $\theta$ unchanged. This makes the finetuning process much faster compared to updating all model weights and allows an increase in the performance of the model on multiple instances without needing multiple copies of the model.

\begin{algorithm}[hbt!]
\caption{Heuristic Finetuning with Active Search}\label{alg:activesearch}
\begin{algorithmic}[1]
\REQUIRE SOPCC problem instance $\rho$, generated heuristic $H_{\theta}$ number of iterations $I$, number of solutions per step $K$, number of samples $S$
\ENSURE Finetuned heuristic $H_{\theta}'$
\STATE $H'_{\theta} \gets \text{copy } H_\theta$ and stop gradient flow to network parameters $\theta$.
\FOR{$i=1$ to $I$}
    \STATE Sample K solutions for $\rho$ using Algorithm \ref{alg:walk} for $i\in\{1,\dots,B\}$
    \STATE Compute $\nabla_{\theta}J(H_{\theta}'\mid\rho_i)$ using Eq.~\eqref{eq:reinforce_baseline} (or Eq.~\eqref{eq:reinforce_population}
    \STATE $H_{\theta} \gets \text{AdamW}(H_{\theta}', -\nabla_{\theta}J)$
\ENDFOR
\RETURN $H_{\theta}$
\end{algorithmic}
\end{algorithm}

\subsection{Neural Architecture for Heuristic Generation}

The parameters $\theta$ compute a heuristic $H_\theta$ from the SOPCC instance $\rho$, which includes the graph instance $G=(V,E)$, start node $v_s$, goal node $v_g$, initial budget $B$, failure constraint $P_f$, and the PDF used to sample edge weights. To ensure that $\theta$ can work for graphs $G$ of any size, we use an encoding neural network, described in the following.
As a preprocessing step, we normalize the reward for every node $v_i$ in $G$ using Min-Max Normalization. To incorporate the stochastic edge weights $d_{ij}$ into the model, we sample $S$ independent edge weights using the given PDF, normalize each sample by $B$, and compute the average over all $S$ normalized samples to get $\bar{d}_{ij}$.
For every node $v_i$ in $G$, we define its corresponding node input features
\[
h^{(0)}_i = \bigl(\text{normalized reward } R(v_i), \quad \mathbf{1}_{v_i = v_s}, \quad \mathbf{1}_{v_i = v_g} \bigr),
\]
where $\mathbf{1}_{\cdot}$ denotes a one-hot indicator vector. For every edge connecting node $v_i$ to $v_j$, we define its corresponding edge input features
\[
e^{(0)}_{ij} = \bigl(\bar{d}_{ij}, \quad P_f, \quad \mathbf{1}_{v_i = v_s}, \quad \mathbf{1}_{v_j = v_g} \bigr).
\]

We split the neural network parameters $\theta$ into two parts:
\begin{enumerate}
    \item \textbf{Encoder}: Takes as input node features $\mathbf{h}^{(0)}$ and edge features $\mathbf{e}^{(0)}$, and outputs learned embeddings for each edge
    \[
    \mathbf{e}^L_{ij} \in \mathbb{R}^{n \times n \times d_e},
    \]
    where $n$ is the number of nodes in $G$ and $d_e$ is the embedding dimension.
    \item \textbf{Decoder}: Takes as input the learned edge embeddings $\mathbf{e}^L$ and outputs the generated heuristic $H_\theta$.
\end{enumerate}

To implement the encoder, we use the Edge-Augmented Transformer (EGT) \cite{hussain2022EGT}, which adapts the Transformer architecture introduced in \cite{vaswani2017attention} to jointly learn both node and edge embeddings for a graph. 
At every layer $l$, the EGT layer uses a custom, multi-head, self-attention mechanism that incorporates the node embeddings $\mathbf{h}^{l}$ and edge embeddings $\mathbf{e}^{l}$ into a single computed attention matrix. It then uses this attention matrix to update both the node embeddings and edge embeddings into $\mathbf{h}^{l+1}$ and $\mathbf{e}^{l+1}$, respectively. We refer the reader to the original paper \cite{hussain2022EGT} for a more detailed explanation of the model architecture.
The decoder is a single feedforward network that transforms the learned edge embeddings into the matrix
$
H \in \mathbb{R}^{n \times n}
$
defining the heuristic function $H$.

\section{Results} \label{sec:results}

In this section, we evaluate the performance of our training algorithm to learn heuristics for solving SOPCC instances with varying constraints. We then assess the solution construction using our learned heuristics and test the performance of the active search algorithm on real-world graph instances.
In all experiments, we initialize our model with a node embedding size $d_h = 64$, edge embedding size $d_e = 16$, 4 EGT attention heads, and 3 EGT layers. The total parameter count of our model is 108,458 learnable parameters.
As a benchmark, we use the MILP formulation for SOPCC from \cite{MILP} to compare against our N(CO)$^2$ algorithm. For all MILP runs, we use the GUROBI solver. Each edge is assigned 100 samples. We set a maximum time limit $t_{\max}$ based on the number of nodes $n$ as follows: 
for $n\in[1, 99]$, $t_{max}=600\text{s}$; for $n\in[100,199]$, $t_{max}=900\text{s}$; for $n\in[200,299]$, $t_{max}=1200\text{s}$. 
Following the guidance of the original authors, we lower the failure probability constraint $P_f$ from $[0.1, 0.05]$ to $[0.05, 0.01]$.
As a control, we also run Algorithm~\ref{alg:walk} using a heuristic derived from our previous work \cite{CarpinTASE2024}, which we refer to as the \textit{baseline heuristic}. The baseline heuristic for each edge from $v_i$ to $v_j$ is computed as
\[
\frac{R(v_j)}{\mathrm{mean}(\text{sCosts}(v_i, v_j, S))}.
\]
During solution construction, actions are selected greedily based on the baseline heuristic for each valid edge.
To stay consistent with our previous work, we use the following exponential distribution as the PDF for sampling edge weights:
\[
\kappa d_{ij} + \mathcal{E}\left(\frac{1}{(1-\kappa) d_{ij}}\right),
\]
where $d_{ij}$ is the Euclidean distance from node $v_i$ to $v_j$, and $\mathcal{E}(\lambda)$ denotes a sample from the exponential distribution with rate $\lambda$. We set $\kappa = 0.5$ for all experiments.
We implement the model and training pipeline using PyTorch. We provide a batch-compatible implementation of Algorithm~\ref{alg:walk} written in PyTorch with TensorDict \cite{bou2023torchrl} that maximizes CPU or GPU usage.\footnote{All the code needed to reproduce this paper is available at \url{https://github.com/ucmercedrobotics/sop-heuristic-opt}.}
All experiments were run on a MacBook Pro M2 with 12 CPU cores and 32 GB of RAM. MILP runs were performed with GUROBI utilizing all 12 CPU cores.

\subsection{Training Setup}\label{subsec:trainsetup}

For all training experiments, we train the model for 100 epochs, with 50 steps per epoch and 32 graph instances per step. Training is performed exclusively on graphs with 50 nodes. For each training graph \( G \), the position \((x,y)\) of each node is sampled uniformly at random from the interval \([0,1]\). Node rewards are also sampled uniformly from \([0,1]\). The start node \( v_s \) and goal node \( v_g \) are chosen randomly, allowing the possibility that \( v_s = v_g \). Challenging budgets \( B_i \) are sampled uniformly from \([1.5, 3]\), and the failure constraint \( P_f \) is randomly selected from \([0.01, 0.15]\) in increments of 0.01. For each edge, we generate \( N=100 \) samples of edge weights to be used in heuristic generation and solution construction within Algorithm \ref{alg:walk}. The model is optimized using the AdamW optimizer with a learning rate of 0.001.



\begin{figure}[htb]
    \centering
    \includegraphics[width=246pt]{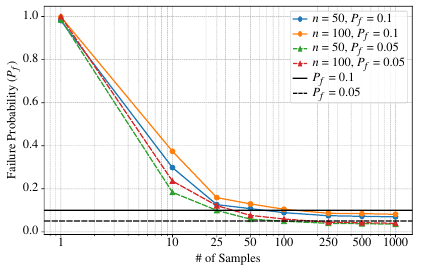}
    \caption{Evaluation of the compliance of paths sampled using Algorithm \ref{alg:walk} with respect to the failure constraint $P_f$ with increasing number of sampled edge costs.}
    \label{fig:samples}
\end{figure}

\begin{table*}[t]
\centering
\begin{tabular}{|c|c|c||c|c|c||c|c|c||c|c|c|} 
 \hline
   &  &  & \multicolumn{3}{c||}{Baseline Heuristic} & \multicolumn{3}{c||}{N(CO)$^2$+AS Heuristic} & \multicolumn{3}{c|}{MILP} \\
 \hline
 Test Case & Budget & $P_f$ & $R$ & $F$ & $t$(s) & $R$ & $F$ & $t_{model}+t_{as}+t_{path}=t$(s) & $R$ & $F$ & $t$(s) \\ [0.5ex] 
 \hline
 berlin52 & 3771 & 0.1 &
 12.81 & 9\% & 0.04 &
 16.42 & 8\% & $(0.004)+(2.95)+(0.06)=3.00$ &
 16.57 & 10\% & 56 \\
 berlin52 & 3771 & 0.05 &
 12.24 & 5\% & 0.04 &
 14.98 & 5\% & $(0.004)+(3.08)+(0.06)=3.14$ &
 15.91 & 6\% & 22 \\
 st70 & 337 & 0.1 &
 13.30 & 7\% & 0.06 &
 16.91 & 10\% & $(0.005)+(4.12)+(0.08)=4.20$ &
 16.54 & 14\% & 216 \\
 st70 & 337 & 0.05 &
 12.52 & 5\% & 0.05 &
 16.51 & 5\% & $(0.005)+(4.03)+(0.08)=4.11$ &
 15.93 & 7\% & 333 \\
 eil101 & 314 & 0.1 &
 22.43 & 11\% & 0.08 &
 32.23 & 10\% & $(0.007)+(6.18)+(0.13)=6.31$ &
 32.99 & 13\% & 736 \\
 eil101 & 314 & 0.05 &
 21.53 & 6\% & 0.07 &
 31.44 & 5\% & $(0.008)+(6.73)+(0.13)=6.87$ &
 32.51 & 8\% & 326 \\
 ch150 & 3264 & 0.1 &
 35.39 & 6\% & 0.11 &
 42.23 & 8\% & $(0.014)+(11.33)+(0.21)=11.55$ &
 43.11 & 16\% & 452 \\
 ch150 & 3264 & 0.05 &
 34.91 & 3\% & 0.11 &
 41.71 & 5\% & $(0.014)+(11.41)+(0.22)=11.64$ &
 43.85 & 11\% & 738 \\
 tsp225 & 1958 & 0.1 &
 54.79 & 10\% & 0.18 &
 69.79 & 11\% & $(0.025)+(31.04)+(0.29)=31.35$ &
 61.01 & 11\% & 704 \\
 tsp225 & 1958 & 0.05 &
 53.86 & 6\% & 0.18 &
 68.29 & 6\% & $(0.028)+(31.44)+(0.29)=31.44$ &
 61.46 & 10\% & 370 \\
 a280 & 1289 & 0.1 &
 63.21 & 10\% & 0.23 &
 79.43 & 9\% & $(0.044)+(57.09)+(0.56)=57.69$ &
 62.25 & 9\% & 836 \\
 a280 & 1289 & 0.05 &
 62.41 & 5\% & 0.26 &
 75.93 & 5\% & $(0.055)+(61.26)+(0.58)=61.90$ &
 58.22 & 5\% & 939 \\
 \hline
\end{tabular}
\caption{Comparison between the different algorithms on the benchmark graphs from \url{http://comopt.ifi.uni-heidelberg.de/software/TSPLIB95/}. The elapsed time for N(CO)$^2$+AS is divided into $t_{model}$ (model inference time to predict initial heuristic), $t_{as}$ (time elapsed for 100 iterations of active search finetuning of heuristic \ref{alg:activesearch}), and $t_{path}$ (time taken to sample path with algorithm \ref{alg:walk} using finetuned heuristic).}
\label{table:results}
\end{table*}

\begin{figure*}[tb!]
    \centering
    \includegraphics[width=500pt]{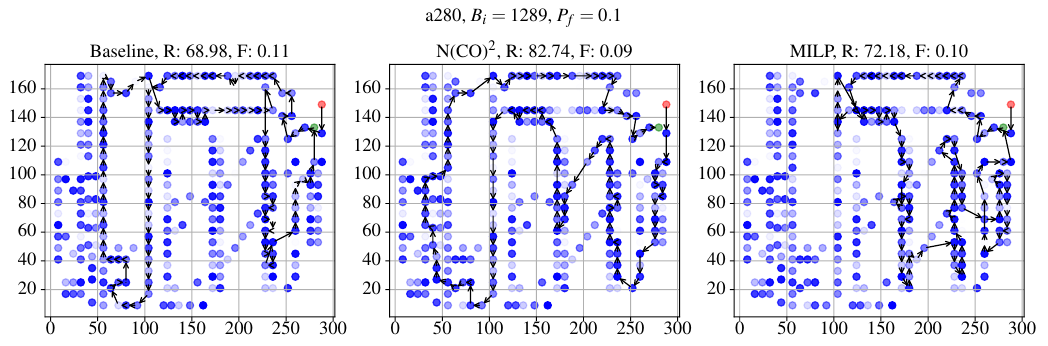}
    \caption{a280 TSP benchmark graph visual of Baseline, N(CO)$^2$ and MILP solution paths with associated rewards and failures. The red node is the start while the green node is the goal. For each shown path, $R$ represents the total reward of the path, while $F$ is the estimated failure probability of the traversing the path.}
    \label{fig:a280}
\end{figure*}

\subsection{Evaluation Datasets}

We created four evaluation datasets to validate the performance of our learned heuristic during the training process. The datasets are as follows: 50 graphs with $n=50, B_i=1.5, P_f=0.1$, 50 graphs with $n=50, B_i=2, P_f=0.05$, 25 graphs with $n=100, B_i=1.5, P_f=0.1$, and 25 graphs with $n=100, B_i=2, P_f=0.05$. The graphs are created using the same setup defined in Section~\ref{subsec:trainsetup}. The two sets of $n=50$ graphs show the performance of our model on the graph size found in training, while the $n=100$ graphs showcase the model's performance on larger graphs that the model has not seen. As a reference, we generate paths for all datasets using both the baseline heuristic and the MILP. We greedily construct baseline paths by running Algorithm~\ref{alg:walk} with the baseline heuristic $K=100$ times per graph and logging the path with the highest reward. We run the MILP once on each graph and log the path found after the set max time limit.

\subsection{N(CO)$^2$ Performance over Training Run}

Figure~\ref{fig:eval} compares the training performance of N(CO)$^2$ using both REINFORCE with baseline \eqref{eq:reinforce_baseline}, and population-based REINFORCE defined in \eqref{eq:reinforce_population}, denoted as REINFORCE$_b$ and REINFORCE$_{pop}$ respectively. At each epoch, both models sample 100 solutions for all four evaluation datasets and log the path with the highest reward. The solutions generated from REINFORCE$_{pop}$ consistently result in a higher reward compared to those generated using REINFORCE$_b$.

\subsection{Failure Probability of Solution Construction with Increasing Sample Size}

We show that given a sufficient number of sample costs $N$ in Algorithm~\ref{alg:walk}, the failure probability of the constructed path approaches the chance constraint $P_f$ (Figure~\ref{fig:samples}.) We use the heuristic $H_{\theta_{pop}}$ generated from the fully trained model using \eqref{eq:reinforce_population} for all runs. To estimate the failure probability of each sampled path $\mathcal{P}$, we sample 10,000 costs for each edge in $\mathcal{P}$ and compare the average total cost of the path against the initial budget $B_i$. Using $S$ between 100 and 250 ensures that the failure probability  is close to the value of $P_f$.

\subsection{Performance of Active Search-based Finetuning on Larger Graphs}
\begin{figure}[htb]
    \centering
    \includegraphics[width=246pt]{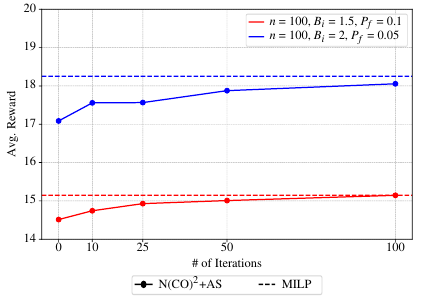}
    \caption{Evaluation of active search algorithm on 2 evaluation datasets not seen during pretraining. The reward is averaged across all graph instances.}
    \label{fig:as}
\end{figure}

We evaluate the benefit of using Algorithm~\ref{alg:activesearch} to finetune the generated heuristic $H_\theta$ on individual graph instances. 
In Figure~\ref{fig:as}, we run the finetuning algorithm with a varying number of iterations on the 100-node evaluation datasets to show the benefit of active search on graph instances not seen during training. 
For each graph instance, we use the generated heuristic $H_{\theta_{pop}}$ from the fully trained model using \eqref{eq:reinforce_population} as the initial heuristic. 
We finetune using $K=100$ iterations, then sample 100 solutions with Algorithm~\ref{alg:walk} using the updated heuristic, and log the path with the highest reward. We can see in Figure~\ref{fig:as} that increasing the number of finetuning steps directly correlates with better performance of the heuristic.

\subsection{Performance on Real-World Graphs}

Following our previous work \cite{CarpinTASE2024}, we borrow benchmark graphs from TSPLIB \cite{TSPLIB} to evaluate the performance of our model on real-world graphs. These graphs consist of various sizes and topologies, so to compute a challenging initial budget $B_i$ for each instance, we divide the cost of the optimal path given by TSPLIB by a factor of 2. We compare the results averaged over 50 runs using the baseline heuristic, 50 runs using our trained model with active search using \eqref{eq:reinforce_population}, and 5 runs of the MILP. For each run of the baseline heuristic, we construct 100 solutions and log the path with the highest reward. For each run using our model, we generate the initial learned heuristic, and then finetune the heuristic on each graph instance using Algorithm~\ref{alg:activesearch} with $K=100$. For graphs with less than 200 nodes, we use 50 iterations of active search, and for graphs with more than 200 nodes, we use 100 iterations of active search. For each run using the MILP, we run the program for the max time and log the best found solution and the initial time taken to find that solution. Table~\ref{table:results} shows the results for each algorithm.

On smaller graph instances, the performance of the MILP is better than our algorithm since the search space is smaller. However, as the number of nodes increases past 200, the performance of the MILP scales much worse than our algorithm. For tsp225 and a280, our algorithm is able to surpass the solution of the MILP algorithm given a max time limit. The MILP will be able to find a better solution than our model given enough time, but the time required to reach that solution might not be worth the cost if we can construct a similar quality solution in much less time. Figure~\ref{fig:a280} shows the paths with the best reward found by all algorithms for the a280 graph instance.



\vspace{-3mm}
\section{Conclusions and Future Work} \label{sec:conclusions}
In this paper, we have presented a novel framework for learning heuristics to solve problems with chance constraints. Algorithms based on heuristics, such as MCTS, are commonly used for solving complex optimization problems in automation, and their performance is highly influenced by the quality of the heuristic. Traditionally, heuristics are manually designed; however, our contribution demonstrates that they can be learned from data, even when considering constrained problems. Our experimental validation supports our conclusion that this is a viable approach.
In the future, we aim to expand this framework to address other problems with stochastic constraints that arise in automation.

\bibliographystyle{plain}
\bibliography{report.bib}

\end{document}